\documentclass{article}

\usepackage{microtype}
\usepackage{graphicx}
\usepackage{subfigure}
\usepackage{booktabs} % for professional tables

\usepackage{color}
\usepackage{natbib}
\usepackage{booktabs}
\usepackage{float}
\usepackage{graphicx}
\usepackage{multirow}
\usepackage{amsthm}
\usepackage{amsmath}
\theoremstyle{definition}
\newtheorem{definition}{Definition}
\DeclareMathOperator{\piscore}{\mathbf{\pi}-score}
\DeclareMathOperator{\piscoreb}{\mathbf{\pi}-\textbf{score}}
\newcommand{\bs}{\boldsymbol}
\DeclareMathOperator*{\argmin}{argmin}

\usepackage{hyperref}

\usepackage[accepted]{icml2021}

\icmltitlerunning{Partially Interpretable Estimators (PIE): Black-Box-Refined Interpretable Machine Learning }

\begin{document}

\twocolumn[
\icmltitle{Partially Interpretable Estimators (PIE): Black-Box-Refined Interpretable Machine Learning }

% It is OKAY to include author information, even for blind
% submissions: the style file will automatically remove it for you
% unless you've provided the [accepted] option to the icml2021
% package.

% List of affiliations: The first argument should be a (short)
% identifier you will use later to specify author affiliations
% Academic affiliations should list Department, University, City, Region, Country
% Industry affiliations should list Company, City, Region, Country

% You can specify symbols, otherwise they are numbered in order.
% Ideally, you should not use this facility. Affiliations will be numbered
% in order of appearance and this is the preferred way.
% \icmlsetsymbol{equal}{*}

\begin{icmlauthorlist}
\icmlauthor{Tong Wang}{uiowa}
\icmlauthor{Jingyi Yang}{uiowa}
\icmlauthor{Yunyi Li}{goo}
\icmlauthor{Boxiang Wang}{uiowa}
\end{icmlauthorlist}

\icmlaffiliation{uiowa}{University of Iowa}
\icmlaffiliation{goo}{University of Texas at Austin}

\icmlcorrespondingauthor{Tong Wang}{tong-wang@uiowa.edu}
\icmlcorrespondingauthor{Bo Xiang Wang}{boxiang-wang@uiowa.edu}

% You may provide any keywords that you
% find helpful for describing your paper; these are used to populate
% the "keywords" metadata in the PDF but will not be shown in the document
\icmlkeywords{Machine Learning, ICML}

\vskip 0.3in
]

% this must go after the closing bracket ] following \twocolumn[ ...

% This command actually creates the footnote in the first column
% listing the affiliations and the copyright notice.
% The command takes one argument, which is text to display at the start of the footnote.
% The \icmlEqualContribution command is standard text for equal contribution.
% Remove it (just {}) if you do not need this facility.

%\printAffiliationsAndNotice{}  % leave blank if no need to mention equal contribution
\printAffiliationsAndNotice{\icmlEqualContribution} % otherwise use the standard text.

\begin{abstract}
% Feature attribution methods have been widely used for explaining black-box models, where each feature is assigned a contribution score for a prediction. Despite the simple form of these methods, many of them are post hoc explanations, which have been criticized for non-perfect fidelity, inconsistency, etc. In this paper, 
We propose Partially Interpretable Estimators (PIE)  which attribute a prediction to individual features via an interpretable model, while a (possibly) small part of the PIE prediction is attributed to the interaction of features via a black-box model, with the goal to boost the predictive performance while maintaining interpretability.  As such, the interpretable model captures the main contributions of features, and the black-box model attempts to complement the interpretable piece by capturing the ``nuances'' of feature interactions as a refinement. 
We design an iterative training algorithm to jointly train the two types of models. Experimental results show that PIE is highly competitive to black-box models while outperforming interpretable baselines. In addition, the understandability of PIE is comparable to simple linear models as validated via a human evaluation. 
\end{abstract}

\section{Introduction}
% \textcolor{red}{positioning the paper: over-think? black-box too complicated? and then interpretability}

% 
Model interpretability has been increasingly desired in various real-world applications of machine learning in recent years. The need for interpretability comes from both model creators and model users - model creators wish to understand the inner workings of a model in order to validate, debug, and improve the model, and model users wish to understand how a prediction is generated in order to trust and adopt the model. 

Existing solutions to achieving interpretability can be categorized into two broad areas. The first is to apply explainer methods that provide post-hoc explanations for a given black-box model. However, some concerns have been brought up  \citep{rudin2019stop,laugel2019dangers} on potential issues of black-box explainers. For example, there may exist inconsistency \citep{ross2017right} in the explanations since there is no ground truth and different explanations could be generated for the same prediction, by different explainers or same explainer with different parameters; the explanations can also be intentionally or unintentionally be deceptive \citep{aivodji2019fairwashing}.  
These issues result from the fact that the explainers, after all, are not the decision-making process themselves.
Therefore, the second area is receiving much attention in recent years, that focuses on building models that are inherently interpretable, such as rule-based models, decision trees, linear models, case-based models, etc., which do not need external explainers. While interpretable models have been advanced to achieve very competitive performance compared to black-box models in some examples by extensively searching the model space, interpretable models still fail often in the competition, given the constraint in model structure and complexity being simple and easy to understand.

In this paper, we propose a new framework for tabular data, where we \textbf{refine} the predictions of an interpretable model with a black-box model that is jointly trained with it. This collaboration of two types of models aims to preserve the understandability and intelligibility of interpretable models while utilizing the strength of a black-box mode to boost predictive performance.  We desire two properties when building such a model. First,  the interpretable model should be very simple such that it can be understandable for any technical and non-technical users. In this paper, we adopt the additive feature attribution mechanism as the interpretable piece.  Second, the contribution of the black-box should be as small as possible such that it only refines the prediction instead of dominating it, which will largely preserve the partial interpretability.  %We call it Partially Interpretable Estimators (PIE).

% In light of the motivations above, we aim to design an inherently interpretable model which adopts the feature attribution mechanism for its understandability. 
The feature attribution mechanism is widely adopted for prediction and explanation. Representative works include SHAP \citep{lundberg2017unified} and its variants \citep{lundberg2018consistent,shrikumar2017learning,schwab2019cxplain}, which assign contributions to each feature used in a model, and the sum of the contributions is the difference between a model's output and the target mean. For example, when a model predicts the income of a person to be \$140,000 while the average income is \$60,000, SHAP decomposes their difference, \$80,000, to \$40,000  for having a Ph.D. degree, \$20,000 for majoring in computer science, and \$20,000 for the site being in California. This form of the model is of particular interest to researchers and practitioners since it is very easy to understand and directly shows how much a feature contributes to a prediction. 
Feature attribution mechanism is also adopted in predictive models such as linear models and generalized additive models (GAM), where a prediction can be decomposed into contributions of each feature. However, these models are often inferior to black-box models (if no feature engineering or interactions are included) due to the simplistic model form and linearly additive nature of features.
%Here, we propose an approach to address this limitation while preserving the interpretability. 

To compensate the limitation in performance, we refine the prediction with a black-box model.
Our model is built based on a key hypothesis that a prediction may not be perfectly decomposed to individual features, \emph{so we allow a possibly small part of the prediction to be attributed to the interaction of features}. The idea is that the individual features can capture the main trend in data, ignoring the ``nuances'' in a prediction, while the interaction of features refines the interpretable piece, generated by a black-box model. In this paper, we use additive models to capture the main contributions from individual features and gradient boosting tree for the non-interpretable piece. The two models are trained jointly to globally optimize the predictive performance. % It can be considered as ``amendment'' to the interpretable prediction.
 Then, as long as the non-interpretable refinement, which we regularize during training, only accounts for a small proportion in the final output, we can consider the prediction largely interpretable. We name the model Partially Interpretable Estimator (PIE). 

%Therefore, the interpretable piece provides the main insights in the features,  while the non-interpretable piece amends the interpretable prediction with the goal to increase the accuracy.
 %Two types of solutions have been proposed to gain interpretability. One is to build self-explaining models such as linear models and decision trees, which do not rely on other methods to provide post-hoc explanations. The other is to build explaining methods which extract insights from black-box models to assist human understanding. The former is often criticized for the downgrade in predictive performance because the model is built under the objective of being simple and small, in addition to being accurate. The latter, black-box explainers, are also under heated debate \cite{rudin2019stop,aivodji2019fairwashing} about whether they truly reflect the synergies and relations of features inside the black-box they try to explain.

% 
Then, for the same example above, our model may predict the income of a person to be \$150,000 because this person has a Ph.D. degree, which contributes \$40,000, majors in computer science, which contributes \$80,000, works in California, which contributes another \$20,000. Then the black-box model produces an additional \$10,000, considering the interactions of features that cannot be disentangled to individual features. 

% The relationship between the white and black space is analogous to the white and black keys in a PIE, where white keys create a bone structure of a melody while the black keys make some critical decorations which may affect the overall style of a musical piece. Therefore, we name our model PIE - Postfixing Interpretable Archetype with orthogonal NOn-interpretability. 

The evaluation of PIE models considers both predictive performance and interpretability. Since a prediction comes from an interpretable model and a black-box model, larger \emph{contribution} from the interpretable model is desired for interpretability purposes. We represent the interpretable contribution as the ratio of the explained variation (i.e., $R^2$) by the interpretable piece only to that of also using the black-box model as refinement, i.e., the percentage of explained variation that is due to the interpretable model only. We coin this metric $\pi$-\textit{score}. We test our model on public datasets.  We compare our model with three sets of baselines: (1) interpretable baselines ($\pi$-score of one) such as linear models and generalized additive models, (2) partially interpretable baseline, EBM ($\pi$-score less than one), and (3) a black-box baseline ($\pi$-score of zero), XGBoost. %We call the predictive performance - interpretable contribution plot \emph{interpretable contribution map}.

Finally, we validate the understandability of PIE models via a survey evaluation with 57 human subjects, where we compare PIE with and linear models, when PIE has different $\piscore$s. Results show the PIE models are very comparable to linear models in terms of  ease of understanding.

% The rest of the paper is organized as follows. In Section \ref{sec:related}, we review existing work related to PIE and discuss the distinctions. Section \ref{sec:pie} presents the PIE model and Section \ref{sec:train} describes the training algorithm. Experimental evaluations are shown in Section \ref{sec:exp} and the human evaluation is presented in Section \ref{sec:human}. Section \ref{sec:conclusion} discusses the conclusions. 

\section{Related Work}\label{sec:related}
We will first review GAM with interaction terms and then discuss other models of forms similar to ours.

% additive; Neural interactions; SHAPLEY;LIME
% GAM with interactions
% Hybrid

% \paragraph{Black-Box Explainers} Our work is related to but different from black-box explainers. The explainers explain a black-box model  locally \cite{ribeiro2016should} or globally \cite{adler2016auditing, lakkaraju2017interpretable}, providing some insights into the black-box model by identifying key features, interactions of features \cite{tsang2017detecting}, etc. %The explanations are extracted from post hoc approximations which provide some insights into the black-box model by identifying key features or interactions of features \cite{tsang2017detecting}. 
% One representative work is LIME \cite{ribeiro2016should} that explains the predictions of any classifier by learning an interpretable linear model locally around the prediction. Our work is different in that the interpretable model directly participates in the predictive performance, therefore guaranteeing 100\% fidelity on the part it produces. The GAM model does not explain the neural network, but instead, explains the how individual features affect the output in the data.
\noindent\textbf{Generalized Additive Models and Their Extensions} GAMs \citep{HT90} combine single-feature models called shape functions through a linear function. The shape functions can be arbitrarily complex. Canonical shape functions are based on various forms of splines \citep{marx1998direct,HT90}, trees \citep{lou2013accurate}, or neural networks in a recent work  Neural Additive Models (NAM) \citep{agarwal2020neural}. The additive nature of traditional GAMs has been the biggest advantage in interpretability and, meanwhile, a limitation for not considering feature interactions. The ignorance of interactions of features causes a possible loss in the predictive performance. Therefore, some earlier work proposed to add selected interactions into the model. For example, \citet{coull2001simple} incorporates factor-by-curve interactions which consider the interactions between numeric features and categorical features. $\text{GA}^2$M \citep{lou2013accurate, caruana2015intelligible} adds pair-wise interaction terms for all features, which improves the overall predictive performance at a complexity cost of $\Omega(d^2)$ extra terms ($d$ is the number of features). We tackle this problem by using one black-box term to capture the interactions among all features. We will compare with a classic GAM, an improved version of $\text{GA}^2$M, called Explainable Boosted Machine (EBM) and the most recent NAM in this paper.% These networks are trained jointly and can learn arbitrarily complex relationships between their input feature and the output. We will compare with NAM for the predictive performance.

\noindent\textbf{Model Hybrids } The idea of utilizing an interpretable model and a black-box model jointly to make a prediction is similar to recent works of hybrid models \citep{wang2019gaining,pan2020interpretable}. The main difference is that the hybrid models partition the feature space, such that an instance is predicted by one of the models, either the interpretable model or the black-box model. Our proposed PIE models, however, partition the \emph{output space}, where each prediction comes from both an interpretable piece and a non-interpretable piece. The idea is also similar to another recent work \citep{sani2020semiparametric} that builds a simple parametric model using a set of pre-defined interpretable features and then adds a more complicated function consisting of uninterpretable features. Unlike PIE that is trained iteratively till convergence, their model is constructed in one pass - first building an interpretable model and then builds black-box model. In addition, their solution works only for regression while PIE also works for classification.

\noindent\textbf{Feature Attribution as Black-box Explanations}
Feature attribution is a type of method that assigns a contribution/importance score to each feature for a specific input \citep{sundararajan2017axiomatic,shrikumar2017learning}. SHAP \citep{lundberg2017unified} computes contributions of a feature based on classic shapley value estimation from cooperative game theory. It assigns a value to each feature that represents its effect on the model's output. The sum of the contributions describes how much a prediction deviates from the mean. Some other attribution methods produce gradient explanation, where an importance score for a feature is a function of the gradient around the input. These methods include Integraded Gradients \citep{sundararajan2017axiomatic}, SmoothGrad \citep{smilkov2017smoothgrad}, Guided GradCAM \citep{selvaraju2016grad}, etc. Motivated by gradient explanation, DeepLIFT \citep{shrikumar2017learning} propose to use slope to describes how y changes as x differs from the baseline.
The key difference between PIE and the above methods is that PIE is a predictive model by itself, not a post-hoc explainer, so the attribution faithfully represents how a prediction is produced and therefore does not suffer from any complaints related to explainer methods \citep{rudin2019stop}. 
% \paragraph{Disentangling Feature Interactions}
% Neural Interaction Transparency (NIT) \cite{tsang2018neural} disentangles the shared learning across different interactions to obtain their intrinsic lower-order and interpretable structure. This is done through a novel regularizer that directly penalizes interaction order. 

\section{Partially Interpretable Estimator}\label{sec:pie}
We work with a training data $\mathcal{D} = \{(\mathbf{x}_i, y_i)\}_{i=1}^n$, where $\mathbf{x}_i = (x_{i0}, x_{i1}, \ldots, x_{ij}, \ldots, x_{id}) \in \mathcal{R}^{d+1}$, 
$y \in \mathcal{R}$ for regression and $y \in \{-1, 1\}$ for classification. Note that PIE works for tabular data. Without loss of generality, here for each $\mathbf{x}_i$ we include $x_{i0} = 1$ to account for the intercept. 
% Consider a regression problem $y = f(\mathbf{x}) + \epsilon$, where $\mathbf{x} = (x_1, \ldots, x_j, \ldots, x_p) \in \mathcal{R}^p$, $y \in \mathcal{R}$, and $\epsilon$ is the error term. 
In this work, we consider fitting a model $f$, whose output can be decomposed into an interpretable piece, which is a sum of contributions of individual features, as well as an uninterpretable piece, which represents the interaction of features. When we present the model below, we will ignore the instance index $i$ unless it is necessary.

\begin{definition}
Partially Interpretable Estimator, i.e., \textbf{PIE}, is a predictive model whose prediction equals the sum of contributions of individual features and interactions of features via a link function.
\end{definition}
PIE can be used for regression and classification. For example, if the link function is an identity function, PIE is a regressor; if the link function is a sigmoid function, then PIE becomes a classifier.
\begin{definition}
Given an input $\mathbf{x}$, the contribution of feature $x_j$ to the prediction $f(\mathbf{x})$ from a PIE is called \textbf{pie value} of feature $x_j$, written as $\pi_j(x_{j})$; the contribution of interaction of features is CRoss-feature Uninterpretable coST, i.e., \textbf{crust value},  written as $\kappa(\mathbf{x})$. 
\end{definition}\vspace{-1mm}
The crust value $\kappa(\mathbf{x})$ cannot be written as a sum of contributions to individual features. 

Accordingly, an output of PIE model $f$ can be gleaned as
\begin{equation}
    f(\mathbf{x}) = \sum_{j=0}^d \pi_j(x_j) + \kappa(\mathbf{x}).
    \label{eq:f_pi_kappa}
\end{equation}
We aim to employ two types of models, $g$ for estimating the interpretable piece (pie values) and $\kappa$ for estimating the uninterpretable piece (crust value). Here, denote
\begin{equation}
    g(x) = \sum_{j=0}^d \pi_j(x_j).
\end{equation}%The pie values can be estimated via an linearly additive model such as a linear model or a generalized additive model. The crust value can be obtained via a black-box model that can be learned via gradients. In this paper we use tree ensembles. We elaborate both models below

\subsection{Pie Value Model}
The pie values can be modeled via an additive structure, such as a generalized additive model. Thus, for each coordinate $x_j$, $j = 1, \ldots, d$, we consider a linear basis expansion, 
\begin{equation}
\begin{aligned}
\pi_j(x_j) &= \sum_{k=1}^K {\alpha}_{jk}\psi_{jk}(x_j),
\end{aligned}
\label{eq:w_j}
\end{equation}
$K$ is the number of basis functions and $\pi_0(x_0) = \alpha_0$ corresponds to the intercept. 

The popular choices of the basis function $\psi_{jk}$ include B-splines and cubic splines; the details can be found in \cite{HT90}, for example. The estimation of function $g(\cdot)$ thereby boils down to estimating the basis coefficients $\alpha_{jk}$. 
 
\paragraph{Sparsity} In model~\eqref{eq:w_j}, despite its simple additive structure, the model may still be hard to interpret if $g(\mathbf{x})$ embraces an excessive number of features. To this end, we assume that some pie values are zero, and $g(\mathbf{x})$ involves only a subset of the features. Such sparse representation also comforts with a common scientific hypothesis that the target variable may be affected by only a few important features. We will discuss later how to use a regularization term in the learning objective to identify zero pie values and thus find the set of active features.

%\begin{equation*}
%\begin{aligned}
%g(\mathbf{x}) 
%&= {g}_0+ \sum_{j\in \mathcal{S}} {g}_{j}(x_{j}) \\
% &= \sum_{j\in \mathcal{S}}^p \sum_{k=1}^K {\alpha}_{jk}\psi_{jk}(x_j)\\
%&= \sum_{j\in \mathcal{S}}^p \bar{\bs\psi}_j(\mathbf{x})' {\bs\alpha}_j,
%\end{aligned}
%\end{equation*}
%where $\mathcal{S}$ is set of the active features and will be determined by a regularization term in the learning objective we will discuss later. 

\paragraph{Evaluating $\piscoreb$} Since the overall explainability of a prediction is desired, one would like a prediction to be contributed more to the individual features than interaction of features. Thus we define a $\piscore$ to represent the overall interpretability of a PIE model. It is defined as
\begin{equation}\label{eqn:piscore}
    \piscore(f, \mathcal{D}) = \frac{\text{R}^2(g, \mathcal{D})}{ \text{R}^2(f, \mathcal{D})}.
\end{equation}
where $\text{R}^2$ is the coefficient of determination, and $\text{R}^2$ measures how much variation in the target variable can be explained by a model. Therefore, a $\piscore$ represents \emph{the percentage of variation explained by the interpretable model $g$}. 
A large $\piscore$ indicates that the majority in the target variation can be explained by the interpretable piece (pie values), thus highly interpretable. On the other hand, a small $\piscore$ implies that $\kappa$ plays a significant role in achieving high predictive performance and meanwhile, the PIE prediction is less interpretable. % The $\piscore$ will be used later in the experiment section to evaluate how understandable a PIE prediction is. 

\subsection{Crust Value Model}
In PIE, the uninterpretable piece $\kappa$ is devised to capture the synergies and patterns that cannot be captured by the interpretable piece ${g}$. To estimate $\kappa$, in this work, we propose to employ gradient boosting, which ensembles $T$ regression trees, 
\vspace{-1mm}
\begin{equation}
{\kappa}(\mathbf{x}) = \sum_{t=1}^T {\beta}_{(t)}(\mathbf{x}).
\end{equation}
Each $\beta_{(t)}$ is a regression tree with $L_t$ leaves and $\beta_{(t)}(\mathbf{x}) = w_{jt}$ if $\mathbf{x}$ is allocated to the $j$th leaf, which is denoted by $\mathbf{x} \in \mathrm{leaf}_j^t$. Thus we have
\begin{equation}
\begin{aligned}
{\beta}_{(t)}(\mathbf{x})=\sum_{j=1}^{L_{t}} {w}_{jt} {I}(\mathbf{x} \in \mathrm{leaf}_j^t),
\label{eq:tree_t}
\end{aligned}
\end{equation}
where ${I}$ is the indicator function.

Boosting trees with a maximum depth of $m$ induces a model with $(m-1)$th-order interactions \citep{FHT00}. The introduction of gradient boosting hence captures the interactions and creates an ``refinement'' to $g$ to enhance the predictive performance. 

Besides gradient boosting, one can broaden the choices by using other gradient-based black-box methods such as neural nets.

\subsection{Learning Objective}
Both models $g$ and $\kappa$ are jointly trained via a global objective,
\begin{equation}
\begin{aligned}
\min_{g, \kappa} \dfrac{1}{n}\sum_{i=1}^n L\left (y_i, g(\mathbf{x}_i) + \kappa(\mathbf{x}_i)\right) + \lambda_1 J_1(g) + \lambda_2 J_2(\kappa),
\label{eq:pie_estimation}
\end{aligned}
\end{equation}
where $L$ is a smooth convex loss function with common choices including least-squared loss $L(y,f)=(y -f)^2$ for regression and logistic loss $L(y,f)=\log(1 + \exp(-yf))$ for classification. 

Two penalty terms $J_1(g)$ and $J_2(\kappa)$ are imposed in problem~\eqref{eq:pie_estimation} and $\lambda_1$ and $\lambda_2$ are the corresponding tuning parameters. We now discuss those two penalties.

The penalty $J_1(g)$ is chosen to be the grouped lasso penalty \citep{YL06},
\begin{equation}
\begin{aligned}
J_1(g) = \sum_{j=1}^d \Vert \bs\alpha_j \Vert_2, 
\label{eq:J1}
\end{aligned}
\end{equation}
where each $\Vert \bs\alpha_j \Vert_2 = \sqrt{\sum_{k=1}^K \alpha_{jk}^2}$ denotes the $\ell_2$ norm. The penalty $J_1(g)$ is used to select a set of active features $S \subset \{1, 2, \ldots, d\}$. For any $j \notin S$, the whole group of $\alpha_{jk}$ is discarded and thus the corresponding pie value $\pi_j(x_j) = 0$.

The penalty $J_2(\kappa)$ is imposed as a tree-structured regularization: \vspace{-1mm}
\begin{equation}
J_{2}(\kappa) = \sum_{t=1}^T \mathcal{J}_{2t} (\beta_{(t)}),
\end{equation}
where for each tree, the number of leaves and the leaf values are bounded: 
\vspace{-1mm}
\begin{equation}
\mathcal{J}_{2t} (\beta_{(t)}) = \sum_{j=1}^{L_t} \left(1 + w_{jt}^2 \right).
\end{equation}
The above tree-structured regularization has been introduced by \cite{JT13} to prevent the boosting machines from overfitting. In addition, penalizing $J_{2}(\kappa)$ also regularizes the contribution from the interactions, which preserves interpretability. % Later we will show how tuning $\lambda_2$ affects predictive performance and interpretability. (it is mentioned in the next paragraph. )

%The two penalties are devised to select active features and regularize boosting machines. 
Notably, simultaneously imposing the two penalties in model~\eqref{eq:pie_estimation} essentially leads to a \textit{tradeoff between interpretability and predictability}. The tradeoff effect is governed by the two tuning parameters, $\lambda_1$ and $\lambda_2$. In an extreme case when $\lambda_1 < \infty$ and $\lambda_2 = \infty$, the uninterpretable piece $\hat{\kappa}$ vanishes, making PIE purely interpretable. When $\lambda_1 = \infty$ and $\lambda_2 < \infty$, PIE fully arises from a black-box method; in such case, the prediction accuracy of PIE is high, as long as the black-box method delivers high accuracy. The tradeoff effect of interpretability and predictability will be demonstrated in the experiments. When solving PIE from problem~\eqref{eq:pie_estimation}, $\lambda_1$ and $\lambda_2$ are selected by cross-validation, 

\section{Training Algorithm}\label{sec:train}
In this section, we develop an algorithm to solve $g$ and ${\kappa}$ in problem~\eqref{eq:pie_estimation}. With a given pair of $(\lambda_1, \lambda_2)$, on the basis of additive splines and gradient boosting, problem~\eqref{eq:model} can be expressed as
\begin{equation}
\begin{aligned}
\min_{g, \kappa}&\dfrac{1}{n}\sum_{i=1}^n L \left(y_i, \alpha_0 + \sum_{j=1}^d \bar{\bs \psi}_j(\mathbf{x}_i)' {\bs\alpha}_j + \sum_{t=1}^T \beta_{(t)}(\mathbf{x}_i) \right) \\
&+ \lambda_1 J_1(g) + \lambda_2 J_2(\kappa),
%&+ \lambda_1  \sum_{j=1}^d \Vert \bs\alpha_j \Vert_2 + \lambda_2 \sum_{t=1}^{T}\left( J_{t}+\sum_{j=1}^{J_{t}} w_{jt}^{2}\right).
\label{eq:model}
\end{aligned}
\end{equation}

We use $\hat{g}^{(t)}$ to collect the solution $\bs\alpha_{j}^{(t)}$ and $\alpha_0^{(t)}$ at step $t$, and we denote by $\hat{\kappa}^{(t)}$ the ensemble of regression trees $\sum_{s=1}^t \hat{\beta}_{(s)}(\mathbf{x})$, where each tree is specified in \eqref{eq:tree_t}.

Denote by $Q(g; \kappa)$ the first term in the objective function in \eqref{eq:model}. First initialize $\hat{g}^{(0)} = \hat{\kappa}^{(0)} = 0$. In the $t$-th step, $t=1, 2, \ldots, T$, we fix $\kappa$ at $\hat{\kappa}^{(t-1)}$ and problem~\eqref{eq:model} becomes
\begin{equation}
\begin{aligned}
\min_{g}&\dfrac{1}{n}\sum_{i=1}^n L \left(y_i, a_0 + \sum_{j=1}^d \bar{\bs \psi}_j(\mathbf{x}_i)' {\bs\alpha}_j + \hat{\kappa}^{(t-1)} \right) + \lambda_1 J_1(g).
\label{eq:opt_g}
\end{aligned}
\end{equation}
Because of the nonsmooth of the grouped lasso penalty term, we develop an algorithm based on the proximal gradient descent \citep{PB14}. In the $t$-th step, we update $\hat{g}^{(t)}$ from $\hat{g}^{(t-1)}$ by conducting a cycle of Gauss-Seidel blockwise coordinate proximal gradient descents. 

In a blockwise manner, suppose we have updated $\hat{\bs\alpha}_1^{(t)}, \ldots, \hat{\bs\alpha}_{j-1}^{(t)}$, and we now update $\hat{\bs\alpha}_{j}^{(t)}$. 

Define a proximal operator for the grouped lasso:
$$
\mathrm{prox}_{\lambda_1\delta\Vert \bs\alpha \Vert_2} (\bs \gamma) = \argmin_{\bs\alpha} \left[\lambda_1 \Vert \bs\alpha\Vert_2 + \dfrac{1}{2\delta}\Vert \bs\alpha - \bs\gamma \Vert_2^2 \right],
$$
where $\delta$ is a step size to be determined. The proximal gradient descent algorithm updates $\hat{\bs\alpha}_{j}^{(t)}$ by
\begin{equation}
\begin{aligned}
\hat{\bs\alpha}_j^{(t)} = \mathrm{prox}_{\lambda_1\delta_j \Vert \bs\alpha \Vert_2} \left(\hat{\bs\alpha}_{j-1}^{(t)} - \bs\nabla_j\right),
\label{eq:PGD}
\end{aligned}
\end{equation}
where $\bs\nabla_j$ is a $K$-vector
$$
\bs\nabla_j = \dfrac{dQ(g; \kappa)}{d\bs\alpha_j}\bigg|_{g = \tilde{g}, \kappa=\hat{\kappa}^{(t-1)}}
$$
and the gradient is computed at the current solution $\hat{\kappa}^{(t-1)}$ and
$$
\tilde{g} = \left\{\hat{\bs\alpha}_1^{(t)}, \ldots, \hat{\bs\alpha}_{j-1}^{(t)}, \hat{\bs\alpha}_{j}^{(t-1)}, \ldots, \hat{\bs\alpha}_{d}^{(t-1)}, \hat{\alpha}_0^{(t-1)}\right\}.
$$
Solving problem~\eqref{eq:PGD} is equivalent to solving
\begin{equation}
\begin{aligned}
\min_{\bs\alpha_j \in \mathcal{R}^K} \;\; & Q(\tilde{g}, \hat{\kappa}^{(t-1)}) + \bs\nabla_j'\left(\bs\alpha_j - \bs\alpha_j^{(t)}\right) \\
&+ \dfrac{\delta_j}{2} \Vert \bs\alpha_j - \bs\alpha_j^{(t)}\Vert_2^2 + \lambda_1 J_1(\tilde{g}),
\label{eq:aj_update}
\end{aligned}
\end{equation}
whose solution is 
\begin{equation*}
\begin{aligned}
\hat{\bs\alpha}_j^{(t)} = \left(\hat{\bs\alpha}_j^{(t-1)} - \delta_j \cdot \bs\nabla_j\right) \left(1 - \dfrac{\lambda_1}{\left\Vert\hat{\bs\alpha}_j^{(t-1)} / \delta_j - \bs\nabla_j\right\Vert}_2 \right)_+.
\end{aligned}
\end{equation*}

We now determine the step size $\delta_j$. The goal is to guarantee that the objective of \eqref{eq:opt_g} decreases after updating $\hat{\bs\alpha}_j^{(t)}$ in \eqref{eq:PGD}. To achieve the goal, we need the objective value of \eqref{eq:aj_update} to be no less than that of \eqref{eq:opt_g}, and this can be satisfied if a Lipschitz condition on $L$ holds, i.e., $|L'(u_1) - L'(u_2)| < M |u_i - u_2|$. We have $M=1$ for the least-squared loss and $M=1/4$ for logistic loss. We then choose the step size $\delta_j$ to be the largest eigenvalue of $\frac{1}{nM}\bs\Psi_j'\bs\Psi_j$, where $\bs\Psi_j$ is an $n \times K$ matrix whose $i$th row is $\bar{\bs \psi}_j(\mathbf{x}_i)$.

In the $t$-th step, after all $\hat{\bs\alpha}_j^{(t)}, j = 1, \ldots, d$, have been computed, we update $\hat{\alpha}_0^{(t)}$. Since the grouped lasso penalty is not imposed on the intercept, the update of $\hat{\alpha}_0^{(t)}$ is from solving a smooth problem. We apply one step of gradient descent:
\begin{equation*}
\begin{aligned}
\hat{\alpha}_0^{(t)} = \hat{\alpha}_0^{(t-1)} - \delta_0\frac{dQ(g; \hat{\kappa}^{(t-1)})}{d \alpha_{0}}\bigg|_{g = \tilde{g}, \kappa=\hat{\kappa}^{(t-1)}},
\end{aligned}
\end{equation*}
where $\tilde{g}=\{\hat{\bs\alpha}_1^{(t)}, \ldots, \hat{\bs\alpha}_d^{(t)}, \hat{\alpha}_0^{(t-1)}, \hat{\kappa}^{(t-1)}\}$ is the current fit. Likewise, to ensure that the objective value of problem~\eqref{eq:opt_g} decreases after updating $\hat{\alpha}_0^{(t)}$, we let the step size $\delta_0 = 1/n$ for regression and $1/(4n)$ for logistic regression.

So far we have updated $\hat{g}^{(t)}$ in the $t$-th step and we now update $\hat{\kappa}^{(t)}$. The key idea is to reduce the objective in problem~\eqref{eq:model} by adding the current solution a regression tree that fits the negative gradient. We compute the pointwise negative gradient of $L$ at the current solution: 
$$
r_i^t = -\dfrac{dL(y_i, f_i)}{df_i} \bigg|_{f_i = \tilde{f}_i},
$$
where $\tilde{f}_i = \hat{\alpha}_0^{(t)} + \sum_{j=1}^d \bar{\bs \psi}_j(\mathbf{x}_i)' \hat{\bs\alpha}_j^{(t)} + \sum_{s=1}^{t-1} \hat{\beta}_{(s)}(\mathbf{x}_i)$. We then use a regression tree $\hat{b}$ to fit each $r_i^t$,
\begin{equation}
\begin{aligned}
&\argmin_{b}\dfrac{1}{n}\sum_{i=1}^n \left (r^t_i - b(\mathbf{x}_i) \right)^2 + \lambda_2 J_{2t}\left(\beta_{(t)}\right).
%=&\argmin_{\beta}\dfrac{1}{n}\sum_{i=1}^n \left (-r_i \beta_{(t)}(\mathbf{x}_i) + (\beta_{(t)}(\mathbf{x}_i))^2 \right) + \lambda_2 J_{2t}\left(\beta_{(t)}\right).
\label{eq:tree}
\end{aligned}
\end{equation}
In this work, we adopt the top-down ``best-fit" strategy proposed by \cite{FHT00} to build regression trees, while other good algorithms can be used as well. After fitting $\hat{b}$, we let $\hat{\kappa}^{(t)} = \hat{\kappa}^{(t-1)} + \hat{\beta}_{(t)}$, where $\hat{\beta}_{(t)}=\nu\hat{b}$ and $\nu$ is a small constant dubbed the shrinkage parameter that is suggested by \cite{FHT00} to improve the estimation. 

Consequently, we have updated both $\hat{g}^{(t)}$ and $\hat{\kappa}^{(t)}$ in the $t$-th step. We keep iterating the aforementioned procedure until convergence. %, that is, 
%\begin{equation*}
%\begin{aligned}
%\dfrac{Q(\hat{g}^{(t-1)}; \hat{\kappa}^{(t-1)}) - Q(\hat{g}^{(t)}; \hat{\kappa}^{(t)})}{Q(\hat{g}^{(t-1)}; \hat{\kappa}^{(t-1)})} < \epsilon.
%\end{aligned}
%\end{equation*}

%the change of the objective value in problem~\eqref{eq:model} is less than a threshold. 

%For any given tree structure of $\beta_{(t)}$, let $m_{l}^t$ be the number of observations in leaf $l$, problem~\eqref{eq:tree} becomes 
%\begin{equation*}
%\begin{aligned}
%& \argmin_{w} \dfrac{1}{n} \sum_{l=1}^{J_t} \left[ -\left(\sum_{i \in \text{leaf}_l^t} r_i^t w_{lt} \right)  + (m_{l}^t+\lambda_2) w_{lt}^2 + \lambda_2\right],
%\end{aligned}
%\end{equation*}
%which gives rise to $\hat{w}_{lt} = \sum_{i \in \text{leaf}_{l}^t} r_i^t/ (m_{l}^t + \lambda_2)$. We then need to select the tree structure of $\beta_{(t)}$. 

\vspace{-1mm}
\section{Experiments}\label{sec:exp} \vspace{-1mm}
We conduct detailed experimental evaluation of PIE models on public datasets.  We compare with state-of-the-art interpretable, partially interpretable, and black-box baselines.\footnote{We focus on regression problems in the experiments  and readers can refer to the supplementary material for additional results for classification.}
% We evaluate the PIE model on public datasets. We compare with other interpretable and non-interpretable baselines. In addition, we study the tradeoff of interpretability and predictive performance of PIE models. Finally, we investigate the sensivity of the performance to different model parameters.

% First, we evaluate the predictive performance of PIE using public datasets.

\begin{table*}[h]
\centering
\caption{The average and standard deviation of test RPE from 5-fold cross-validation.}\label{tab:regression}\vspace{1mm}
\small
\begin{tabular}{c|c|cc|ccccc}
\toprule
\multirow{2}{*}{\textbf{Datasets}}&\multicolumn{1}{|c|}{\textbf{Black-box}}& \multicolumn{2}{c|}{\textbf{Partially Interpretable}} & \multicolumn{4}{c}{\textbf{Interpretable}} \\ \cline{2-9 }
&
  \textbf{XGBoost} &
  \textbf{EBM} &
  \textbf{PIE} &
  \textbf{PIE-GAM} &
  \textbf{GAM} &
  \textbf{NAM} &
  \textbf{Lasso} &
  \textbf{Ridge} \\ \hline
winequality  & \textbf{.499(.011)} & .543(.005) & .519(.026) & .672(.024)& .675(.020)& .974(.032) & .710(.015) & .710(.015)   \\
CASP        & .361(.012) & .495(.004)& \textbf{.359(.006)} & .691(.016) & .681(.008) & .782(.034) &.718(.008) & .718(.008)   \\
CBM         & \textbf{.000(.000)}  & \textbf{.000(.000)}& \textbf{.000(.000)} &\textbf{.000(.000)}& .001(.000) & .002(.001) &.004(.000) & .004(.000)  \\
gridp        & .046(.003)& .059(.003)& \textbf{.045(.003)} & .218(.009)& .217(.009)& .286(.008) & .354(.012) & .354(.012)    \\
energyp     & \textbf{.420(.011)} & .653(.012)& .448(.012) & .891(.037) & .833(.006)& .856(.017)  & .856(.007) & .856(.007)   \\
parkinsons & \textbf{.002(.000)} &.026(.002) & \textbf{.002(.000)} & .107(.003)  & .076(.005) & .099(.011) & .093(.005) & .093(.005) \\
Crime & .143(.076) & .268(.151) & \textbf{.077(.030)} & .778(.231)& .091(.029)& .540(.130) & .114(.029) & .132(.067)\\
blog & \textbf{.371(.032)} & .771(.146) & .492(.062) & .756(.074)& .595(.019) & .978(.090) & .611(.019)& .611(.019)\\
% traffic & \textbf{.054(.001)} & .466(.121) & .078(.003) & .215(.005)& .219(.006)& &.217(.007) & .217(.007)\\  
\bottomrule
\end{tabular}
\end{table*}
\vspace{-1mm}
\subsection{Predictive Performance} 
\noindent\textbf{Datasets} 
  We use some publicly available datasets at the UCI Machine Learning repository \citep{Dua:2019}. See Table \ref{sec:data} for a summary of the datasets used in the experiments. We pre-process the data by applying one-hot encoding to categorical features and normalizing the numeric features. 
\begin{table}[ht]
\footnotesize
\centering
    \caption{A summary of datasets for evaluation. Denote by ${n}$ and ${d}$ the number of instances and the number of features in each dataset.}\label{sec:data}\vspace{1mm}
\begin{tabular}{@{}l|rrr@{}}
\toprule
\textbf{Datasets}      & $\bs{n}$      & $\bs{d}$       \\ \midrule
% adult         & 30,000 & 15 & classification \\
% creditcard    & 30,000 & 24  & classification \\
% recidivism    & 11,645 & 107 & classification \\
% sqf141516     & 80,754 & 27 & classification \\
% TelcoCustomer & 7,032  & 20  & classification \\
% magic04       & 19,019 & 11  & classification \\
winequality   & 6,497  & 13   \\
CSAP          & 45,730 & 10   \\
CBM           & 11,934 & 14   \\
gridp         & 10,000 & 13   \\
energyp       & 19,735 & 25   \\
parkinsons    & 5,875  & 21   \\ 
crime         & 2,215  & 120\\
blog          & 26,199 & 280\\
% traffic & 40777& 176 \\
\bottomrule
\end{tabular}
\end{table}

\noindent\textbf{Baselines}  We first apply XGBoost \citep{chen2016XGBoost} to the datasets, providing a benchmark to examine how much (if any) predictive performance PIE may lose by enforcing interpretability. We then compare PIE with the following interpretable models: Lasso, Ridge, and GAM, and the most recent Neural Additive Model (NAM) \cite{agarwal2020neural}. All the four models are additive models, and the prediction can be decomposed into contributions from individual features. The comparison demonstrates the gain of PIE in predictive performance by allowing the interaction of features in PIE. We also include a comparison with the Explainable Boosted Machine (EBM), which is a fast derivative of GA2M \citep{caruana2015intelligible}. Like PIE, EBM also handles interactions of features, but EBM includes multiple pairwise interactions which follows heuristics to select the pairs to be included.
As the most relevant competitor, EBM is compared against PIE in the $\piscore$ and predictive performance.  
The model setup and parameter tuning can be found in the supplementary material.
% \paragraph{Setup} We conduct a 5-fold cross-validation for each dataset and each model, where the training data in each fold is further split into 80\% training and 20\% validation for parameter tuning.  For CART, we minimize cost complexity factor. For lasso and ridge, we tune $\lambda$  from $(0.0001,0.001,0.01,0.1,1,10,100)$.  Finally, for XGBoost, we tune nround with max number $10000$ and stop at best nround. For our PIE model, we choose $\lambda_1$ from $(1,0.1,0.01, 0.005, 0.001, 0.0005, 0.0001)$ and $\lambda_2$ from $(0, -1, \ldots, -4)$ on the $\log 10$ scale. The shrinkage parameter is fixed to be 0.05 and the convergence criterion is set to be $10^{-5}$. The models with the best performance for on the validation set for each method are evaluated on the test set. 

\noindent\textbf{Evaluation Metrics } For each method, we evaluate the predictive performance according to the relative prediction error (RPE),
\begin{equation}\label{eqn:rpe}
\text{RPE} = \frac{\sum_i(y_i - \hat{y}_i)^2}{\sum_i(y_i -\bar{y})^2}.
\end{equation}

In addition, we also quantify the contribution of interpretable models via $\piscoreb$ from formula (\ref{eqn:piscore}). Since we desire a prediction that is ``largely'' interpretable, we tend to favor a large $\piscore$.  

\noindent\textbf{Results} Table \ref{tab:regression} summarizes the average and standard deviation of the performance across five folds. PIE is highly competitive to XGBoost, beating all other baselines, including interpretable and partially interpretable competitors. %The high prediction accuracy of PIE is largely due to the predictive power of tree ensembles. 
It is also observed that PIE outperforms EBM. Although both PIE and EBM capture feature interaction, the methods are quite different in how the interactions are estimated. While EBM relies on heuristics to choose pairwise interactoins, PIE uses one term to accommodate all feature interactions and optimize it under a global objective.
%which allows pairwise feature interactions. %The better performance of PIE is partly attributed to the iterative training method we devise under a global learning objective instead of a heuristic-based approach.

A PIE model jointly trains an interpretable model and a black-box model to benefit the overall prediction. We conjecture that the joint training may also benefit the predictive performance of the interpretable model. By capturing the ``nuances'' from the data, the black-box method may force the interpretable model to concentrate on the main trends and thus prevent it from overfitting. To address such conjecture, we extract the solely interpretable part, namely GAM, from the fitted PIE model. We call this procedure \textbf{PIE-GAM}. We evaluate the predictive performance and exhibit the results in Table~\ref{tab:regression}. Surprisingly, we find that PIE-GAM is highly competitive to other interpretable models, even though PIE-GAM is not originally crafted as a predictive method but a collaborator with gradient boosting in the PIE model. The good performance of PIE-GAM implies another practice of PIE. If users do not tolerate any non-interpretability of PIE, \textit{they may choose to simply ignore the black-box part and only use the interpretable piece: the predictive performance may still be competitive to other interpretable baselines.}

\begin{table*}[ht]
\centering
\caption{The average and standard deviation of RPE and $\piscore$s on the test sets from 5-fold cross-validation. The sparse PIE models use not more than 8 non-zero features.  }\label{tab:sparse}\vspace{1mm}
\small 
\begin{tabular}{c|cc|cc|cc}
\toprule
\multirow{2}{*}{\textbf{Datasets}}
& \multicolumn{2}{c|}{\textbf{EBM} }& \multicolumn{2}{c|}{\textbf{PIE}}& \multicolumn{2}{c}{\textbf{Sparse PIE}}\\ \cline{2-7}
&
  \textbf{RPE} &
  \textbf{$\piscore$} &
  \textbf{RPE} &
  \textbf{$\piscore$} &
  \textbf{RPE} &
  \textbf{$\piscore$}\\ \hline
  winequality & .543(.005) & .670(.017) & .519(.026)& .682(.027)& .519(.018)&  .414(.137)\\
  CASP &.495(.004) & .735(.006) & .359(.006) & .482(.025) & .382(.004) & .482(.005)\\
  CBM & .000(.000) & .977(.001) & .000(.000) & 1.00(.000) & .000(.000)& 1.00(.000)\\
  gridp &  .059(.003) & .821(.006) & .045(.003)& .822(.008) &.045(.003) &.331(.002)\\
  energyp & .653(.012) & .788(.019) & .448(.012)  & .196(.065) & .451(.011)& .058(.005)\\
  parkinsons & .026(.002)& .812(.017) & .002(.000)& .895(.003) & .002(.000)& .895(.003)\\
  crime &.268(.151) &.943(.041) & .077(.030) & .235(.238) & .065(.011) & .485(.400)\\
  blog & .771(.146)  & .406(.124)& .492(.062) & .473(.099) & .503(.075)& .278(.046)\\
%   traffic & .466(.121)& \textcolor{red}{score} &  .078(.003) & .851(.006) & .133(.017)& .730(.024) \\
  \bottomrule
\end{tabular}
\end{table*}

\noindent\textbf{$\piscoreb$ Analysis} The $\piscore$s for EBM and PIE are reported in Table 3. We observe that one some datasets (winequality, CBM, gridp, parkinson), PIE models have higher $\piscore$s compared to EBM while achieving smaller RPE, indicating a more efficient trade-off of interpretability and predictive performance. On other datasets, PIE models have smaller $\piscore$s as well as smaller RPEs than EBM, implying that gradient boosting plays a more important role in boosting the predictive performance PIE than the pairwise interaction terms in that of EBM. This explains the better predictive performance of PIE than EBM. Combining Table \ref{tab:regression} and \ref{tab:sparse}, we find that the $\piscore$s are small when interpretable baselines cannot achieve a good performance so that PIE has to rely heavily on gradient boosting to achieve small prediction error. See the energyp dataset as an example: PIE reduces RPE by almost the half as compared to the interpretable baselines, and the $\piscore$ is only 0.2. On the other hand, when the interpretable baseline is capable of obtaining good performance, then gradient boosting has a minimal participation in the prediction, yielding a relatively large $\piscore$. Such case is exemplified by the CBM dataset.

% \textcolor{red}{things to add}
% \begin{itemize}
%     \item \textcolor{red}{add a bseline where we train an interpretable model first and then fit a black-box to the residual? (I got the question every time I talk to people about this idea so it must be answered convincingly)}
%     \item \textcolor{red}{more classification datsets where XGBoost is the best}
% \end{itemize}

% \begin{table*}[h]
% \begin{tabular}{ccccccccc}
% \hline
% \textbf{datasets} &
%   \textbf{Decision Tree} &
%   \textbf{Lasso} &
%   \textbf{Ridge} &
%   \textbf{Gam} &
%   \textbf{XGBoost} &
%   \textbf{PIE} &
%   \textbf{PIE Lasso Only} &
%   \textbf{Interpretability} \\ \hline
% TelcoCustomer & .500(.000) & .587(.014)          & \textbf{.589(.017)} & .586(.014) & .582(.014) & .570(.008)          & .570(.007) & .999 \\
% creditcard    & .641(.015) & .767(.016)          & .768(.016)          & .768(.014) & .772(.020) & \textbf{.771(.015)} & .771(.015) & 1.00 \\
% adult         & .846(.003) & \textbf{.906(.004)} & \textbf{.906(.004)} & .889(.004) & .925(.005) & .905(.003)          & .902(.003) & .997 \\
% sqf141516     & .547(.019) & \textbf{.678(.003)} & \textbf{.678(.003)} & .674(.003) & .695(.012) & .666(.002)          & .500(.000) & .750 \\
% magic04       & .813(.008) & .839(.011)          & .839(.011)          & .896(.005) & .933(.004) & \textbf{.900(.003)} & .896(.004) & .996 \\ \hline
% \end{tabular}
% \end{table*}

% Please add the following required packages to your document preamble:
% \usepackage{booktabs}

\vspace{-1mm}
\subsection{PIE with Sparsity Constraints}\vspace{-1mm}
The models reported in Table \ref{tab:regression} were selected only based on predictive performance. In many real-world applications, the sparsity of the model is highly desired since it is relatively easier to understand a model with fewer non-zero factors contributing to the prediction. We propose a procedure named \textbf{sparse PIE}: we tune $\lambda_1$ and $\lambda_2$ in the PIE model, and we select only among the models with the maximum of {eight non-zero features (so that the total explanation entities is 10, including an interaction term and the overall prediction)}. We then evaluate sparse PIE in terms of RPE and $\piscore$s. Table \ref{tab:sparse} summarizes the results.

It is interesting to find, with the promotion of sparsity, PIE downgrades the performance only minimally, still beating every other baseline except XGBoost. However, PIE relies more on the non-interpretable part since the interpretable model's representability has been discounted by the restriction of no more than eight non-zero features. Therefore, the $\piscore$s decrease for sparse PIE compared to PIE models.

The above result suggests that allowing the interpretable part to be more expressive/complicated (such as using more features) can potentially increase the $\piscore$, because PIE can attribute more contribution to the interpretable part rather than resort to the black-box component. The interaction term steps in when the interpretable part alone is incapable of a good predictive performance. A natural tradeoff thus arises between $\piscore$ and the sparsity of the interpretable model, which needs to be made by users when tuning the model in practice.

% \subsection{Interpretability - Accuracy tradeoff}
% We investigate the tradeoff between interpretability and accuracy of PIE models, by tuning the parameters XX and XX to obtain a set of different models. We choose XX datasets for demonstration and plot the models in Figure \ref{fig:tradeoff}. Results show that ....
% \begin{figure}[ht]
% \centering
% \includegraphics[width=0.5\textwidth]{Experiment_graph/parkinsons/Interp_normal_graph.png}
% \caption{Example of a parametric plot ($\piscore$, RPE )}
% \end{figure}

\vspace{-1mm}
\subsection{Tradeoff and Sensitivity Analysis}\vspace{-1mm}
To understand the effect of the model parameters, we run experiments with different combinations of $\lambda_1$ and $\lambda_2$. We then evaluate RPE and $\piscore$ on test sets. Figure \ref{fig:sensitivity} depicts the results for parkinsons dataset. (More results are shown in the supplementary material.) It shows that RPE is more sensitive to $\lambda_2$, while $\piscore$ is more sensitive to $\lambda_1$. The RPE grows with $\lambda_2$ while $\piscore$ increases as $\lambda_1$ decreases. The findings are consistent with the discussion in Section \ref{sec:pie}. The results also suggest that users are provided with the flexibility to tune the model to obtain smaller or larger $\piscore$, based on the need of the task.
\begin{figure}[h]
\centering
\includegraphics[width=0.39\textwidth]{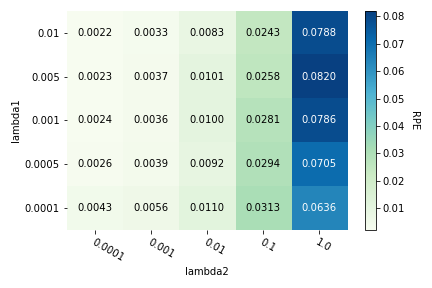}
\includegraphics[width=0.39\textwidth]{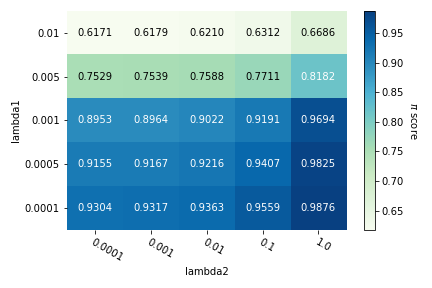}
\vspace{-3mm}
\caption{Sensitivity analysis of $\lambda_1$ and $\lambda_2$.}\label{fig:sensitivity}
\end{figure}

\vspace{-2mm}
\subsection{Case Studies}
\begin{figure}[ht]
\centering
\includegraphics[width=0.49\textwidth]{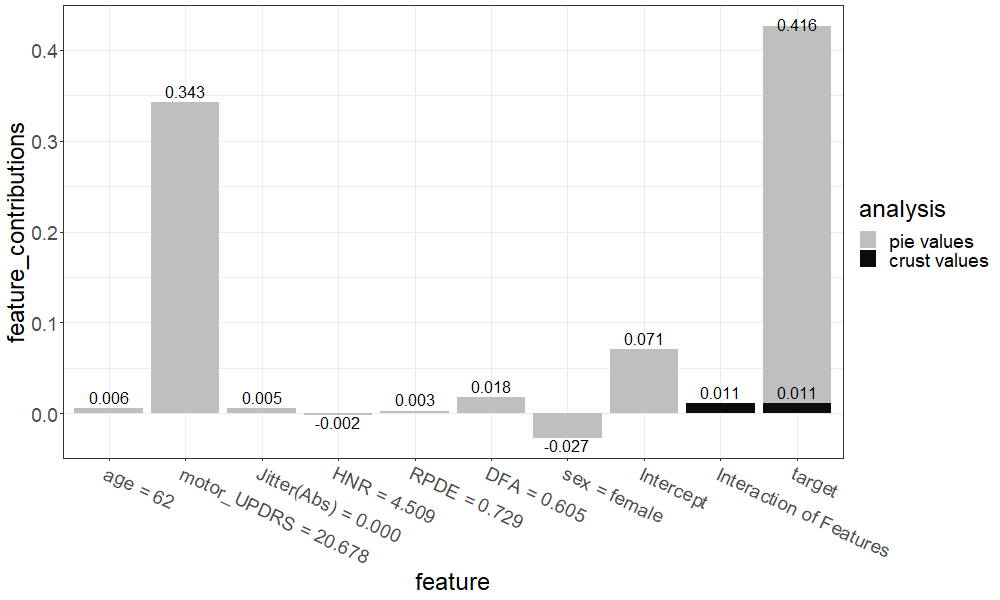}
\vspace{-1mm}
\caption{An example of the prediction breakdown for an instance from the Parkinsons data.}\label{fig:example}
\end{figure}

\vspace{-1mm}
We provide some visualizations of PIE models in this section. We choose an example from the parkinsons dataset. This dataset predicts a patient's Parkinson's disease symptom score. Higher value indicates the patient has more severe parkinson's symptons. We plot the pie values and crust value in Figure \ref{fig:example}. In this example, the feature motor\_UPDRS contributes most to the target variable, followed by the intercept value. Motor\_UPDRS describes the motor UPDRS score, where UPDRS stands for Unified Parkinson's Disease Rating. The crust value in this case only contributes 0.011 to the output. %Jitter(Abs) is a measure of fundamental frequency. HNR measures ratio of noise to tonal components in the voice. RPDE is a dynamical complexity measure. DFA is a signal fractal scaling exponent.
Thus the prediction has a high $\piscore$ of 0.92.

\vspace{-3mm}
\section{Human Evaluation}\label{sec:human}\vspace{-1mm}
We conduct a human evaluation for the understandability of PIE, and specifically, to compare with linear models. To do that, we design a survey where we display three models to each subject, a linear model with seven features, a PIE model with six pie values and crust value that accounts for 10\% of the total prediction, and a model randomly chosen from the following: a PIE model with six pie values, but the crust value accounts for 40\% of the total prediction, and a more complicated linear model with 14 features. We then ask each subject to rate how easy it is to understand and explain the model's prediction on a scale of 1 to 5 (5 being very easy to understand). Finally, we ask subjects to provide comments, especially on the models they did not find easy to understand. The questions are provided in the supplementary material. 

\begin{figure}[ht]
\centering
\includegraphics[width=0.35\textwidth]{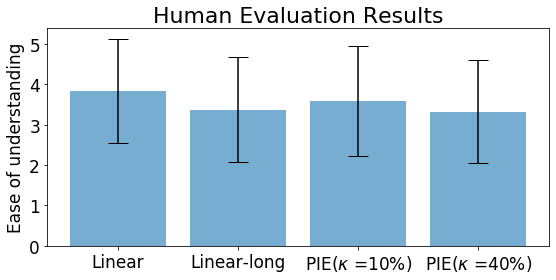}
\vspace{-1mm}
\caption{Ratings of ease of understanding for four models by human subjects. ``Linear'' refers to a linear model with 7 features. ``Linear-long'' refers to the linear model with 14 features.}\label{fig:human}
\end{figure}

A total of 57 subjects took the survey, including undergraduate and graduate students from the computer science department, statistics department, and business schools in an R1 University with an analytics-related major. The average age is 26 and 25 subjects are female. We report their ratings for each model in Figure \ref{fig:human}. 

Results show that (1) since all models are linear or additive, there only exist small differences in the ratings; (2) when both PIE and the linear model have seven items to explain the prediction, PIE is almost as understandable as linear models when the crust value is relatively small (10\% of the predicted value in the survey). Although PIE has a portion of a black-box, it seems that as long as the source of the contribution is identified, people find such an explanation acceptable. Interestingly, this PIE model is more easily to understand than a linear model with 14 features, although the linear model does not have black-box predictions. The result seems to suggest that users are less tolerant with a linear model with more features than a PIE model with fewer features and a relatively small black-box portion; (3) When the crust value increases from 10\% to 40\% of the output, the rating of the ease of understanding decreases. To understand this, we look at the comments on this PIE model. The main reason is summarized as below: the unknown synergy of features in the crust value accounts for too much of the prediction. Thus they find it ``vague and not as trustworthy'' than other models they saw in the survey.
\vspace{-1mm}
\section{Conclusion}\label{sec:conclusion}\vspace{-1mm}
We proposed a partially interpretable estimator that decomposes a prediction into contributions from individual features via an interpretable model and interaction of features via a black-box model. The black-box model captures the ``nuances'' in data, providing an ``refinement'' of  the interpretable feature attribution.  Experiments on publiclly available datasets show that PIE is highly competitive to XGBoost while beating all interpretable and partially interpretable baselines.  Human evaluation results show that PIE is almost equally understandable as a linear model with the same number of non-zero terms. 

As a by-product, the interpretable model, GAM, in PIE achieves competitive performance compared to interpretable baselines. Therefore, it can be used individually as a predictive model if one desires complete transparency. This makes PIE highly adaptable and flexible in real applications.

PIE provides a novel way of attending to the collaboration between an interpretable model and a black-box model, utilizing the strength of the latter in predictive peformance and the former for interpretability.

% In the unusual situation where you want a paper to appear in the
% references without citing it in the main text, use \nocite
\nocite{langley00}

\bibliography{GAM}
\bibliographystyle{icml2021}

%%%%%%%%%%%%%%%%%%%%%%%%%%%%%%%%%%%%%%%%%%%%%%%%%%%%%%%%%%%%%%%%%%%%%%%%%%%%%%%
%%%%%%%%%%%%%%%%%%%%%%%%%%%%%%%%%%%%%%%%%%%%%%%%%%%%%%%%%%%%%%%%%%%%%%%%%%%%%%%
% DELETE THIS PART. DO NOT PLACE CONTENT AFTER THE REFERENCES!
%%%%%%%%%%%%%%%%%%%%%%%%%%%%%%%%%%%%%%%%%%%%%%%%%%%%%%%%%%%%%%%%%%%%%%%%%%%%%%%
%%%%%%%%%%%%%%%%%%%%%%%%%%%%%%%%%%%%%%%%%%%%%%%%%%%%%%%%%%%%%%%%%%%%%%%%%%%%%%%
% \appendix
% \section{Do \emph{not} have an appendix here}

% \textbf{\emph{Do not put content after the references.}}
% %
% Put anything that you might normally include after the references in a separate
% supplementary file.

% We recommend that you build supplementary material in a separate document.
% If you must create one PDF and cut it up, please be careful to use a tool that
% doesn't alter the margins, and that doesn't aggressively rewrite the PDF file.
% pdftk usually works fine. 

% \textbf{Please do not use Apple's preview to cut off supplementary material.} In
% previous years it has altered margins, and created headaches at the camera-ready
% stage. 
% %%%%%%%%%%%%%%%%%%%%%%%%%%%%%%%%%%%%%%%%%%%%%%%%%%%%%%%%%%%%%%%%%%%%%%%%%%%%%%%
% %%%%%%%%%%%%%%%%%%%%%%%%%%%%%%%%%%%%%%%%%%%%%%%%%%%%%%%%%%%%%%%%%%%%%%%%%%%%%%%

\end{document}